\documentclass[11pt]{article}
\usepackage{graphicx}
\usepackage{amssymb}
\usepackage{amscd}
\usepackage{mathrsfs}
\usepackage{longtable,lscape}
\usepackage{amsthm}
\usepackage{amsfonts}
\usepackage{amsmath}
\usepackage{bbm}
\usepackage{float}
\usepackage{url}

\newcommand{\compl}{{\mathbb C}}
\newcommand{\real}{{\mathbb R}}
\newcommand{\captionfonts}{\footnotesize}
\makeatletter  
\long\def\@makecaption#1#2{%
  \vskip\abovecaptionskip
  \sbox\@tempboxa{{\captionfonts #1: #2}}%
  \ifdim \wd\@tempboxa >\hsize
    {\captionfonts #1: #2\par}
  \else
    \hbox to\hsize{\hfil\box\@tempboxa\hfil}%
  \fi
  \vskip\belowcaptionskip}
\makeatother 
\begin{document}
\title{Towards a Quantum World Wide Web\footnote{Submitted to: {\it Theoretical Computer Science}}}
\author{Diederik Aerts$^1$, Jonito Aerts Argu\"elles$^2$, Lester Beltran$^1$, Lyneth Beltran$^1$, \\ Isaac Distrito$^3$, Massimiliano Sassoli de Bianchi$^{1}$, Sandro Sozzo$^{4}$  and Tomas Veloz$^{1,5}$ \vspace{0.5 cm} \\ 
        \normalsize\itshape
        $^1$ Center Leo Apostel for Interdisciplinary Studies, 
         Brussels Free University \\ 
        \normalsize\itshape
         Krijgskundestraat 33, 1160 Brussels, Belgium \\
        \normalsize
        E-Mails: \url{diraerts@vub.ac.be},\\\url{lyneth.benedictinelawcenter@gmail.com}, \\
        \url{msassoli@vub.ac.be}
          \vspace{0.5 cm} \\ 
        \normalsize\itshape
        $^2$ KASK and Conservatory, \\
        \normalsize\itshape
         Jozef Kluyskensstraat 2, 9000 Ghent, Belgium
        \\
        \normalsize
        E-Mail: \url{jonitoarguelles@gmail.com}
		\vspace{0.5 cm} \\ 
        \normalsize\itshape
        $^3$ 2511 Eloriaga Street, \\
         \normalsize\itshape
        Santa Ana, Manila, The Philippines
         \\
        \normalsize
        E-Mail: \url{isaacdis3to@yahoo.com}
	  \vspace{0.5 cm} \\ 
        \normalsize\itshape
        $^4$ School of Management and IQSCS, University of Leicester \\ 
        \normalsize\itshape
         University Road, LE1 7RH Leicester, United Kingdom \\
        \normalsize
        E-Mail: \url{ss831@le.ac.uk} 
        \vspace{0.5 cm} \\ 
        \normalsize\itshape
        $^5$ Instituto de Filosof\'ia y Ciencias de la Complejidad \\ 
        \normalsize\itshape
         Los Alerces 3024, N\~un\~oa, Chile \\
        \normalsize
       and Departamento Ciencias Biolo\'gicas, Facultad Ciencias Biolo\'gicas \\ 
        \normalsize\itshape
         Universidad Andres Bello, 8370146 Santiago, Chile \\
        \normalsize
        E-Mail: \url{tveloz@gmail.com} 
       	\\
              }
\date{}
\maketitle
\begin{abstract}
\noindent
We elaborate a quantum model for the meaning associated with corpora of written documents, like the pages forming the World Wide Web. To that end, we are guided by how physicists constructed quantum theory for microscopic entities, which unlike classical objects cannot be fully represented in our spatial theater. We suggest that a similar construction needs to be carried out by linguists and computational scientists, to capture the full meaning carried by collections of documental entities. More precisely, we show how to associate a quantum-like `entity of meaning' to a `language entity formed by printed documents', considering the latter as the collection of traces that are left by the former, in specific results of search actions that we describe as measurements. In other words, we offer a perspective where a collection of documents, like the Web, is described as the space of manifestation of a more complex entity -- the QWeb -- which is the object of our modeling, drawing its inspiration from previous studies on operational-realistic approaches to quantum physics and quantum modeling of human cognition and decision-making. We emphasize that a consistent QWeb model needs to account for the observed correlations between words appearing in printed documents, e.g., co-occurrences, as the latter would depend on the `meaning connections' existing between the concepts that are associated with these words. In that respect, we show that both `context and interference (quantum) effects' are required to explain the probabilities calculated by counting the relative number of documents containing certain words and co-ocurrrences of words.
\end{abstract}
\medskip
{\bf Keywords}: World Wide Web, Conceptual entities, Quantum structures, Information retrieval

\section{Introduction\label{intro}}
The approach that uses the mathematical formalism of quantum theory in Hilbert space to model complex cognitive processes, such as language, perception, judgment and decision-making, has received substantial confirmation in recent years. Quantum probabilistic models have shown their effectiveness over classical Kolmogorovian (or Bayesian) models in the explanation of several `cognitive fallacies and/or effects', which include membership judgments on natural concepts and their combinations, conjunctive and disjunctive fallacies, disjunction effects and violations of expected utility theory in decision-making (see, {\it e.g.}, 
\cite{a2009a,pb2009,k2010,bpft2011,bb2012,hk2013,pb2013,wbap2013,abgs2013,ags2013,ast2014,asdb2015b} and references therein).

Consequently, quantum-theoretic approaches have been promisingly applied also to computer science and even extended to artificial intelligence. Indeed, on one hand, Hilbert space quantum models have shown distinctive advantages over traditional models, based on classical logic and classical probability theory, in semantic analysis, natural language processing and document retrieval situations (see, {\it e.g.}, 
\cite{Widdows2003,Rijsbergen2004,ac2004,bruzacole2005,HouSong2009,Zuccon2010,Piwowarski2010, Frommholz2010,melucci2012,Houetal2013,Xieatal2015,  melucci2015,Lietal2016,Zhang2016,Wang2016} 
and references therein), and on the other hand, cognitive fallacies, {\it e.g.}, the conjunction fallacy and the Guppy effect, have been recently identified in simple document retrieval experiments, using search engines such as `Google' and `Yahoo', on the World Wide Web (simply referred to as the Web, hereinafter, to be here understood in the limited sense of `the Web of text corpora'.) 
\cite{acds2010,a2011,aabssv2016}.

This success in the application of the formalism of quantum theory in Hilbert space, to represent empirical data collected in cognitive experiments with human participants, naturally leads one to also wonder whether a similar approach could be proposed to model the meaning content of large corpora of written documents -- of which a paradigmatic example would be the pages forming the Web -- to possibly capture those conceptual/semantic aspects which are usually revealed when a human mind enters into contact with said documents, typically when reading them.

Usually, one considers that meaning is created when a mind interacts with a written document, and certainly during such an interaction a `creation aspect' is present. However, the view we want to advocate in this article is that there is also a `meaning content', or `meaning dimension', that can be associated with a collection of documents, which is independent of the minds possibly interacting with them, and which goes beyond the patterns of letters, words and sentences they contain. Indeed, this hidden meaning dimension would only be partially manifest in the document, and would require -- this is our thesis -- a particular mathematical formalism to be faithfully represented. 

Written documents, like webpages, are here assumed to collapse from a conceptual realm also available to human minds, because of their cognitive, language-mediated activities, similarly to how interference patterns collapse in physics laboratories as a consequence of the evolution of quantum entities, which however belong to a non-spatial/non-local `quantum realm', not fully representable within a spatio-temporal theater \cite{as2015}.

And in the same way physicists constructed  quantum theory, to try to capture this hidden (but not for this less real) quantum layer, the same can be done, in principle, about the hidden meaning content associable with a collection of documental entities, like those forming the Web. It is precisely the purpose of the present article to provide a first provisional, but promising, answer about the feasibility of such programme of a genuine quantum modeling of the Web, and similar corpora of written documents. 

For this, we proceed as follows. In Sec.~\ref{operational}, we provide a general operational foundation of the modeling of the meaning content of the Web, considering it as an entity that can be in different states and that is formed by sub-entities that can also be associated with states, and be submitted to experiments.

In Sec.~\ref{Hilbert}, the more specific Hilbertian formalism of standard quantum theory is used to further specify the logic and potential of our modeling. In Sec.~\ref{Context}, we show how to model `context effects' that result from the fact that when we ask a question, this will inevitably alter the state of the conceptual entity under consideration. In Sec.~\ref{Interference}, we come to the central point of our modeling, which consists in using `superposition states' to model `emergence and interference effects', which manifest as effects of overextension and underextension of the experimental probabilities. 

In Sec.~\ref{Collapsing}, we specialize our modeling to the situation where the states of single concepts are represented by characteristic function states, only associated with webpages containing the explicit words indicating these concepts. This allows us to more specifically calculate the interval of probabilistic values that can be generated by pure interference effects. Then, in Sec.~\ref{context and interference}, we analyze the situation where both context and interference effects are present, and show that it allows in principle to model all possible data. In Sec.~\ref{Numerical}, we give a few numerical examples, obtained by performing counts of the webpages containing certain words and combinations of words (using Google as a search engine), taken from Hampton's celebrated experiments on natural concepts \cite{h1988a,h1988b}, showing that both context and interference effects are needed to model them. Finally, in Sec.~\ref{Conclusion}, we offer some concluding remarks.

\section{An operational foundation of the Web} \label{operational}

From now on, for convenience, we only refer to that collection of documents that are the pages of the Web (webpages). However, everything we say in the article will remain valid for other collections of documents. What we are here interested in modeling is the `meaning content' of the Web, through the quantum formalism, and not the human cognitive activity, although it is clear that the latter is responsible of having created the former. 

We start by providing an outline of a possible `operational foundation' of  this `meaning structure' of the Web, not reducible to the `printed patterns of inks' (or sequences of  0 and 1 in computers' memories) characterizing its collection of physical documents, but to be especially understood as the (in a sense, more abstract) meaning carried by these documents. Also, the reality of such `meaning content' will be regarded as being independent of the human minds that have created it, have access to it, and can create further meaning out of it. 

More precisely, our goal is to describe the Web as a `conceptual entity', which can be in different states and can be submitted to different contexts, either deterministic or indeterministic, the latter being associated with `processes of change of state' characterized by well-defined irreducible probabilities. It will then be our task, in Sec.~\ref{Hilbert}, to more specifically represent these states, contexts and probabilities by using the versatility of the mathematical language of quantum theory. 

Thus, we firstly have to make clear the distinction between two kinds of Web: the standard (spatial) Web, made of actual webpages, formed by specific collections of letters and words, printed on paper or encoded in computers' memories, and the `meaning entity' that we can associate with it, formed by concepts existing in different combinations, which is the object of our modeling. This (non-spatial) meaning/conceptual entity, which we will simply call the QWeb ({\it i.e.}, the `Quantum Web'), is of course intimately related to the standard Web, in the same way that a concept, say the concept \emph{Fruits}, is intimately related to the different possible printed words that can be used to indicate it.

Consider the similar situation in physics, for example that of an electron. Before the advent of quantum theory, it was believed that an electron was just a corpuscle, but on closer inspection it was later realized that although an electron can leave corpuscular traces (for instance on a detecting screen), its behavior was not reducible to that of a spatio-temporal (classical) particle, considering that it could also give rise to interference patterns, suggesting a wave-like nature. But even a description in terms of a `wave-particle duality' appeared to be insufficient to capture the full reality of an electron, when for instance combining with other electrons, as genuine multi-dimensional (non-spatial) `wave functions' were required to fully and consistently describe this situation. So, already in physics we are confronted with the problem of distinguishing an entity like an electron, whose reality is not reducible to spatio-temporal phenomena, like waves, particles and fields, and the many ways an electron can concretely manifest, within our spatio-temporal theater, by leaving well-defined and readable traces in our measuring instruments. 

{\it Mutatis mutandis}, it is fundamental to distinguish a `conceptual entity', like the concept {\it Fruits}, from the words, sentences, paragraphs, pieces of texts, etc., that are used to indicate it, concretely appearing in certain webpages. Although this is a fundamental demarcation (for example in semiotics~\cite{semiotics1985}), most models of language do not incorporate a conceptual layer from which language can be constructed, but just focus on formalizing the different relations among the language entities and on the methods for thruth evaluations~\cite{Manning1999} (see~\cite{Lapata2005} for a review of these methods applied to the Web). The language entities, however, are not {\it per se}  `entities of meaning', but rather the traces left by them, within the Web's space of manifestation, which is a space formed by the collection of all the existing webpages, in a given moment of our human history \cite{absv2013}.

Another important aspect to emphasize is that although the QWeb is here viewed as a whole entity, which later we will describe as a quantum entity in a given Hilbert space, it is also a `composite entity', formed by complex combinations of concepts. For instance, the conceptual entity {\it Fruits}, viewed as an individual entity, is certainly part of the QWeb as is the case for the conceptual entity {\it Vegetables}, etc., and in our analysis we will sometimes refer to states as describing the entire QWeb, and other times these same states will be interpreted as referring to more specific individual conceptual entities (or their combinations) forming the latter.\footnote{A concept can partly be understood as a combination of other concepts, which in turn are also combinations of other concepts, and so on, which means that concepts possess a natural recursive-like structure. The `quantum superposition' used in the following of this article, and in earlier works \cite{a2009a,abgs2013,ags2013}, indicates that a concept is also more complex in its relational structure than what its recursive tree-like aspect can indicate. In fact, the quantum formalism, as applied to its modeling, reveals part of this additional structure.}

Thus, on one hand we have the QWeb, a meaning entity, and on the other hand we have the standard (spatial) Web, which is a manifestation, or objectification, of the former, and in the same way the QWeb is formed by concepts, like {\it Fruits} and {\it Vegetables}, and their combinations, the standard Web is formed by printed words and combinations of words, which are like the traces left by the former on the Web's canvass, similarly to how electrons can leave traces on a detection screen. 

Now, a truly innovative aspect of Brussels' operational foundation of cognitive psychology \cite{ass2016b} is the understanding of concepts as `entities in well-defined states', rather than mere `containers of instantiations'. Also, it is considered that the states of conceptual entities can change when they interact with different contexts, some of which will be deterministic, {\it i.e.}, such that the change of state can be predicted in advance, with certainty (for instance when some information is added to the description of a given cognitive situation), and others genuinely indeterministic, typically when the state change is the result of a decision process. 

Our understanding of human concepts as `entities in given states' will be here extended, {\it mutatis mutandis}, to our modeling of the QWeb. More precisely, we will assume that it is always possible to attach to the QWeb, considered as a conceptual entity, well defined states, and we will also assume that the latter can additionally be interpreted as possible states of the concepts (and combinations of concepts) forming the QWeb. For example, let $n$ be the overall number of webpages of the Web (at a given time), and let ${\rm W}_i$ be the $i$-th webpage (assuming they would have been ordered), with $i\in \{1,2, \ldots, n \}$. Then, each webpage ${\rm W}_i$ can be associated with a distinct QWeb state $p_i$. 

Note that a webpage ${\rm W}_i$ is not itself to be understood as `a state', in the same way a trace left by an electron on a detection screen is not as such `an outcome-state' of the electron. However, a spot on the detection screen can be associated with that electron state for which the probability of detecting the electron at that specific spot, when in that specific state, would be equal to $1$, and similarly, a given webpage (a spot on the concrete Web's canvas) can also be associated with a specific QWeb state, to be understood as the outcome-state of a certain `QWeb measurement' (that we will define shortly). In other words, webpages play for the entities of meaning of the QWeb the same role that `instantiations', or `objects', play for concepts in cognitive psychology, or the `traces left on a measuring device' play for micro-physical entities like electrons. 

Now, as mentioned above, a QWeb state $p_i$, associated with the webpage ${\rm W}_i$, can also be interpreted as the state of  other conceptual entities forming the QWeb. Indeed, the `state of a concept' is nothing but a specification of the `reality of that concept in a given meaning context'. A webpage always defines a specific meaning context, resulting from the combination of words it contains, and more precisely the meaning associated with such combination of words. Consider again the concept {\it Fruits}. All webpages containing the word `fruits' can certainly be associated with a state of the concept {\it Fruits}. As a simple example, take the webpage only containing the single word `fruits' (if it doesn't exist yet, we can easily create it). Clearly, such a webpage describes the most neutral possible meaning context for the concept {\it Fruits}, and the associated state -- let us denote it $p_{\it Fruits}$ -- can be understood as representing the most basic aspect of the reality of {\it Fruits}, which we simply call its `ground state'.\footnote{More than one word, or combination of words, are  able to convey a same meaning, like in the case of synonyms. This means that `ground states', and states in general, can be associated with a notion of `meaning degeneracy'; a degeneracy that will be removed when synonymous words are placed in the context of other words.}

Consider then a slightly more elaborated webpage, just containing two words, for example the words `juicy fruits' (again, if such two-word webpage doesn't exist yet, we can easily create it). The `juicy' term now plays the role of a `deterministic context' altering the ground state meaning of the concept {\it Fruits}. Thus, this two-word webpage can now be associated with an `excited state' of {\it Fruits} (but also, of course, with an excited state of {\it Juicy}). Similarly, a webpage made of hundreds of words, still provides a meaning context for {\it Fruits}, although a much more complex one, which can be interpreted as an entire `story about {\it Fruits}', and which can also be associated with one of the countless possible excited states of {\it Fruits}. 

Of course, not all webpages contain the word `fruits', however, this doesn't mean they cannot have some meaning connection with {\it Fruits}, and thus could still be associable with possible states of {\it Fruits}. In other words, not only all webpages can in principle be associated with `states of the QWeb', but also with `states of the conceptual entities forming the QWeb', and this is 
regardless of the fact they may contain or not the specific words indicating the conceptual entity the state of which they would describe. 

As we mentioned already, concepts can play the role of deterministic contexts for other concepts, but not all contexts are deterministic, and measurement processes are typical examples of genuinely indeterministic contexts. If in quantum physics a measurement results from the interaction between a micro-physical entity and a macroscopic apparatus, producing the selection of one among a number of potential outcome-states, in a psychological experiment the apparatus is replaced by the minds of the participants subjected to a conceptual entity in a given initial state (describing the reality of the cognitive situation), having to choose only one among a number of possible answers/decisions, in a way that usually cannot be predicted in advance, not even in principle.

Then, what would be a measurement for the QWeb? Of course, different typologies of measurements can be imagined, but in the present article we will only focus on one. 
It consists in the QWeb interacting with an entity sensitive to its meaning, having the $n$ webpages of the Web stored in its (in principle accessible) memory, as stories, so that the result of the interaction is that one of these stories will be told, with a probability that depends on the initial state of the QWeb. An example of this is a search engine (having the $n$ webpages stored in its indexes) used to retrieve some meaningful information, the QWeb initial state being then an expression of the meaning contained in the retrieval query. 

Now, if the initial state of the QWeb is $p_i$, {\it i.e.}, the state associated with the webpage ${\rm W}_i$, then the `tell a story measurement' can be assumed to be a deterministic process,  
just telling the story contained in the webpage ${\rm W}_i$, with probability equal to $1$. 
But the webpages do not exhaust all the possible states of the QWeb. The collection of webpages is just the tip of the iceberg of the QWeb state space; a tip corresponding to what we have considered to be the more concrete kinds of states, those that can leave traces in our spatio-temporal theater, as specific printable ink patterns. Most QWeb states are  non-spatial states, which cannot be associated with specific webpages, and the selection of a specific story will then result in an indeterministic process. 

A very simple example of initial state is a `uniform state' where each webpage has the same probability ${1\over n}$ of being selected as the outcome of the measurement. In other words, when the QWeb is in a uniform state,  
an arbitrary story will simply be told, 
{\it i.e.}, a story `randomly' selected among all the available stories ({\it i.e.}, webpages). Another simple example are what we might call the `locally uniform states', where only $m\leq n$ webpages have the same (non-zero) probability ${1\over m}$ of being selected as an actual story.

\section{A Hilbert space representation}\label{Hilbert}

We want now to use the Hilbertian formalism of quantum theory to more specifically represent the notions that we have introduced in Sec.~\ref{operational}. By doing so, we will also push a bit further our modeling, proposing a possible form for the states describing `composite concepts', {\it i.e.}, concepts that are formed by the combination of other concepts, in the same way as for instance a physical composite system is formed by the combination of different subsystems. 

Having associated the webpages ${\rm W}_i$ with the states $p_i$, $i\in \{1, \ldots, n \}$, which are the outcome-states of the  
`tell a story measurement', 
we can naturally introduce a $n$-dimensional Hilbert space ${\cal H}$, isomorphic to the vector space $\compl^n$ of all $n$-tuples of complex numbers, and represent the webpages' outcome-states $p_i$ by means of $n$ mutually orthogonal vectors $|e_i\rangle$, $\langle e_i|e_j\rangle =\delta_{ij}$, where $\delta_{ij}$ is the Kronecker delta function and we are using Dirac's ket notation. In other words, we are now associating an orthonormal basis $\{|e_1\rangle,\dots,|e_n\rangle\}$ of ${\cal H}$ to the ensemble of existing webpages.

Let then $A$ be a concept ({\it e.g.,} $A$ = {\it Fruits}). Following the usual quantum-theoretic rules, we can represent any state of $A$ (and more generally, any QWeb state) as a linear combination 
\begin{equation}
|\psi_A\rangle=\sum_{j=1}^n a_je^{i\alpha_j}|e_j\rangle,
\label{psiA}
\end{equation}
with $a_j,\alpha_j\in \real$, $a_j\ge 0$, and $\sum_{j=1}^{n}a_j^{2}=1$. Similarly, if $B$ is another concept ({\it e.g.}, $B$ = {\it Vegetables}), we can also write its possible states as $|\psi_B\rangle=\sum_{j=1}^n b_je^{i\beta_j}|e_j\rangle$, and so on, for all the other possible concepts. Note that contrary to the basis vectors $|e_j\rangle$, associated with the webpages (interpreted as outcome-states), 
we will not generally assume the orthogonality of two vectors $|\psi_A\rangle$ and $|\psi_B\rangle$ describing the states of two different concepts $A$ and $B$.

The complex numbers $a_je^{i\alpha_j}$ appearing in the expansion (\ref{psiA}) can be understood as the generalization of the coefficients appearing in the `term-document matrices' and `term co-occurrence matrices' used in natural language processing techniques, like `latent semantic analysis' (LSA) \cite{Deerwester1990,Landauer1998} and Hyperspace Analogue to Language (HAL)~\cite{HAL1}. More precisely, the coefficients $a_j$ can be understood as expressing the meaning connection\footnote{One can provide a more specific interpretation for the normalized weights $a_j$, characterizing the linear combination (\ref{psiA}), by introducing a notion of `quantum meaning bond of a concept $A$ with respect to another concept $W_j$', thus generalizing the `meaning bond' notion introduced in \cite{acds2010,a2011,aabssv2016}. 
In this case, $W_j$ will be considered as the concept associated with the combination of words appearing in the webpage ${\rm W}_j$. However, a comprehensive discussion of the possible quantum generalizations of the notion of `meaning bond' would lead us  beyond the scope of this article, so we plan to return to this issue in a future work.} between the concept $A$ and the meaning content of the webpage ${\rm W}_j$. On the other hand, the role of the phases $\alpha_j$, as we shall see, is that of carrying a portion of the information about possible emergence and interference effects, when concepts in given states are combined, giving birth to meanings that cannot be deduced from those of the individual composing concepts.

It is important to note, however, that the Hilbert space representation we introduce here is fundamentally different from LSA and HAL, and also from the various quantum `latent semantic' approaches developed in the recent years for information retrieval~\cite{Zuccon2010,Piwowarski2010,Frommholz2010,QLSA2011} and natural language processing~\cite{Huertas2009,Blacoe2014,Coecke2015}. 
Namely, a `latent semantic' approach aims at a geometric or probabilistic representation of the meaning of words of a certain language. Such representation is built from the statistics of occurrence and co-occurrence of words computed from a large collection of documents. Next, this theory is used to build representations of the documents in the collection, and the geometric and/or probabilistic structure associated with the words is used to identify word-word, word-document, and document-document semantic relations~\cite{melucci2015,Landauer1998}. In our approach, instead, the idea is to represent the meaning of the `conceptual entities {\it per se}', which encapsulate at a more fundamental level the meaning carried by the printed words and documents.

In particular, we assume that the entire sequence of words that form a document is a collapsed state of a conceptual entity that corresponds to the idea that the creators of the document had in mind when writing it. Similarly, paragraphs, sentences, and words of a document are also assumed to be more concrete forms of the collapse of this concept into the spatio-temporal realm of the document. Thus, the `reconstruction of concepts' from texts can be in principle obtained by performing various measurements on the structure of words in the corpora of texts, following an inverse problem methodology~\cite{absv2013}, in an analogous way to how it is done for quantum states in physics laboratories~\cite{dariano2004}. The geometric and probabilistic structure associated with the concepts can then be used to identify concept-concept semantic relations, which in turn can also be applied to derive word-word, word-document, and document-document semantic relations. 

There are also essential differences of mathematical nature between our approach and the traditional semantic space approaches first introduced in \cite{Saltonetal1975}, 
built upon
real vector spaces, for which the elements of the word-document matrix are interpreted as weights for the appearance of the different words in the different documents \cite{Deerwester1990,Landauer1998,HAL1}. 
In our approach, built upon complex vector spaces, it is instead the square of the absolute values of the complex elements of the word-document matrix that plays the role of weight for the different word-document 
(and, more generally, concept-document) 
meaning connections
(the complex coefficients $a_je^{i\alpha_j}$, in the expansion (\ref{psiA})).
Hence, in our approach the linearity works differently as it is the case in traditional semantic spaces, namely on the level of the complex numbers whose square roots of their moduli are the weights.

\subsection{Context effects}
\label{Context}

In Sec.~\ref{operational}, we have introduced the 
`tell a story' QWeb measurements.
When measurements are performed, context effects should in principle also be considered. These effects can have different origins and logics. Generally speaking, they can be understood as preparations, {\it i.e.}, as (non-destructive) processes that can bring the state of the measured entity in a pre-measurement state, more appropriate, or meaningful, for the measurement to be performed. In physics, a preparation consists in the entity being submitted to some specific constraints that can select a given state, before allowing it to interact with the measuring apparatus. As a typical example, think of the passage of a photon through a filter, to select a specific polarization state. More generally, a physical entity always needs, to be measured, to be brought into a `spatial state of proximity with the measuring apparatus', to allow both systems to interact and therefore produce an outcome.

In human cognition, consider the example of a person who is asked to tell a story. For this to happen, a portion of the person's memory needs to be evocated and become accessible, from which an actual story can then be selected. Although this activation of a portion of the person's memory appears to be more akin to a change of state to be attributed to the measuring apparatus, it can certainly be accounted for as a preliminary change of the state describing the cognitive situation, because only a portion of the totality of the memorized stories will be effectively accessible during the `tell a story measurement'. The situation is similar if, instead of a person, it were 
a search instrument 
that is asked to tell a story: 
depending on the instrument that is used to provide an outcome, certain pages will be accessible and others not, and again this can be understood as a general context effect that can be taken into account by altering in some way the QWeb's initial state. 

Thus, if $|\psi\rangle$ represents the initial QWeb state,  
and the QWeb entity is subjected to the `tell a story measurement', 
we will now assume that this produces a preliminary change of initial state $|\psi\rangle\to |\psi'\rangle$, before the actual (indeterministic) process of selection of a story occurs. Mathematically speaking, we have of course different possibilities for modeling this effect. Here however we shall adopt the simplest one, consisting in using an orthogonal projection operator $N$ ($N=N^\dagger$ and $N^2=N$), so that we have: $|\psi'\rangle = {N|\psi\rangle \over \|N|\psi\rangle\|}$. Coming back to our previous description, we can understand for instance $N$ as the projection operator onto the subspace generated by the stories that are actually available to be retrieved during the measurement. 

So far we have just considered the situation where the 
`tell a story measurement' is about 
a generic story, without any predetermined content. However, 
a measurement can also be about telling  
a more specific story, say a `story about $X$'. When we ask a question of this kind, we must consider that this more specific interrogative context could also alter the state of the Web. If this were the only `context effect', then $N$ would simply reduce to the projection operator $M_X$ onto the subspace of QWeb states that are `states of $X$' (note that a `state of $X$' is not in general also a `story about $X$', the latter being a state that can be associated with an existing webpage). In what follows, however, we will not make any specific hypothesis regarding the nature of the `context effect' that would be present, limiting ourselves to considering the idealization that it can be modeled by means of a projection $N$. 

We want to also consider the possibility of testing the different QWeb states. More precisely, we want to know if certain states are more or less meaning connected to  certain stories. Consider for instance $|\psi_A\rangle$, here assumed to represent a `state of $A$'. We want to test such state to know if it can be also understood as a `story about $X$', with $X$ a given concept, indicated by the printed term ``X'' (for instance, $A$ could be the concept {\it Fruits}, $X$ the concept {\it Olive}, and therefore `X' the printed word ``olive''). In other words, we want to know if concept $A$, in state represented by $|\psi_A\rangle$, is meaning connected to concept $X$, when the latter is in a concrete state, defined by one of the webpages that are `stories of $X$'. 

For this, we introduce the projection operator $M^s=\sum_{i\in I_X} |e_i\rangle\langle e_i|$, where $I_X$ denotes the set of indexes associated with the webpages that are `stories about $X$', and the superscript ``$s$'' indicates that we are considering the more specific subspace of `states of $X$' that is generated by the `stories about 
$X$'.\footnote{The projection operator $M^s$ should be more precisely denoted $M^s_X$, making explicit that it projects onto the subspace of stories about the concept $X$. However, since there are no risks of confusion, we have dropped the $X$-index , not to have a too heavy notation.}
So, when  
the measurement has to do with 
a `story about $X$', and the initial 
QWeb 
state is represented by $|\psi_A\rangle$, we first have a `context effect' $|\psi_A\rangle\to |\psi'_A\rangle$, producing the pre-measurement state vector $|\psi'_A\rangle = {N|\psi_A\rangle \over \|N|\psi_A\rangle\|}$. 
Then, according to the Born rule, the probability $\mu(A)$, with which $|\psi'_A\rangle$ is evaluated to be a `story about $X$', is given by the average 
$\mu(A) = \langle\psi'_A| M^s |\psi'_A\rangle$. So, if $A$ is a concept in state represented by $|\psi_A\rangle$, 
we have:
\begin{equation}
\mu(A) =  {\langle\psi_A | N^\dagger M^s N|\psi_A\rangle\over \|N|\psi_A\rangle\|^2}={\langle\psi_A | N M^s N|\psi_A\rangle\over \langle\psi_A | N |\psi_A\rangle},
\label{muAandmuB}
\end{equation}
where we have used $\|N|\psi_A\rangle\|^2=\langle \psi_A |N^\dagger N|\psi_A\rangle=\langle \psi_A |N^2|\psi_A\rangle=\langle\psi_A | N |\psi_A\rangle$.
Considering again the previous example where $A$ = {\it Fruits} and $X$ = {\it Olive}, $\mu(A)$ can be interpreted as the probability with which a `story about \emph{Olive}' is considered to be meaning connected with 
the concept $A$  (for instance, because such story would well represent $A$),  where the latter is understood as a `meaning entity' in state 
represented by $|\psi_A\rangle$.

\subsection{Combinations of concepts and superposition states}\label{Interference}

We want now to introduce a procedure to represent the states of a concept $AB$ which is the `combination' of two concepts $A$ and $B$. For the two 
concepts {\it Fruits} and {\it Vegetables}, such a combination could be the concept {\it Fruits and Vegetables}. But also the concept {\it Fruits or Vegetables} is a combination of the concepts {\it Fruits} and {\it Vegetables}. For the concepts {\it Animal}, {\it Eat} and {\it Food}, the sentence {\it The Animal Eats the Food} is again a concept which is a combination. The foregoing examples of combinations of concepts are straightforward cases of how 
they  appear in human language. We have studied all three of them in detail within our quantum cognition approach to human language, in earlier work \cite{a2009a,abgs2013,ags2013,a2007,aerts2009,aerts2010a,aabbssv2016}. We however want to generalize the notion of `combination of concepts', and not just have it indicate what it usually is meant to indicate in human language, namely `concepts whose printed form stand in a sentence one following the other'. For example, also when the printed forms of concepts $A$ and $B$ `co-occur' in one and the same webpage we want to consider this as a trace of  the  `combination' of $A$ and $B$. 

There are several reasons for this more general way of considering the notion of `combination'. First, we believe that `co-occurrence' of `printed concepts' on the Web is the most primitive way in which the `emergence of meaning' on the QWeb can be easily identified, by looking at the `physical Web of printed documents'. This emergence of meaning on the QWeb is considered in our approach as due to the `combination' of the considered concepts, where the co-occurrence is precisely the trace left by the combination on the (physical) Web. It is in this sense not a coincidence from our perspective that co-occurrence of words plays such an important role also in traditional semantic space approaches.

Second, in our analogy between quantum particles and concepts, and our inspiration from physics to build a quantum model for the Web, we want to connect the notion of `combination' in the realm of concepts with that of `composition' in the realm of quantum entities. Composition in quantum mechanics, when the 
quantum entities remain identifiable within the composite entity, is described by means of the tensor product of the Hilbert spaces of the sub-entities. Although also this tensor product procedure is applicable to the situation of the combination of concepts, as we have shown in our quantum modeling of data from cognitive experiments  \cite{entanglement2011,entanglement2014}, it is not this type of combination we focus on in the present article. The composition we consider here is the one where quantum situations interfere in a coherent way to make emerge another quantum situation. It is the `superposition principle' of quantum mechanics that is used to describe this type of composition, 
which in our quantum cognition approach was used to account for the emergence of meaning when concepts are combined \cite{a2009a,abgs2013,ags2013,a2007}.    

The typical analogy in physics is the paradigmatic situation of the two-slit experiment, where an interference pattern emerges when the two slits are both kept open, which cannot be deduced from the patterns obtained when only one of the 
two slits is alternatively kept open, but nonetheless can be described by introducing a state that is the superposition of the two one-slit states. So, we also assume that a `state of $AB$', i.e., a `state of the combination of the two concepts $A$ and $B$', 
can be generally represented by a superposition vector: 
\begin{equation}
|\psi_{AB}\rangle= {|\psi_A\rangle + |\psi_B\rangle \over \| |\psi_A\rangle + |\psi_B\rangle\|},
\label{superposition}
\end{equation}
where $|\psi_A\rangle$ and $|\psi_B\rangle$ represent states of $A$ and $B$, respectively.
Note that in (\ref{superposition}) it is implicitly assumed that the two concepts $A$ and $B$ are on a same footing in the combination $AB$, so that the two one-concept states have the same weight in the superposition. Also, $|\psi_A\rangle$ and $|\psi_B\rangle$ are not assumed here to be necessarily mutually orthogonal. 

We repeat that if we would express our meaning connection between the concept $A$ and the webpage ${\rm W}_i$ in a very simple way, namely by means of the `occurrence' of the term ${\rm A}$ in the printed text of the webpage ${\rm W}_i$, and do the same for the meaning connection between the concept $B$ and the webpage ${\rm W}_i$, then a possible `combination of $A$ and $B$' would be expressed by the co-occurrence of the terms ${\rm A}$ and ${\rm B}$ in the printed text of the webpage ${\rm W}_i$. Hence, our procedure of superposition can be considered as a generalisation of the notion of `co-occurrence' as used in traditional semantic spaces approaches. We will work out in detail the situation of combination of concepts when 
the meaning connections between concepts and webpages take the simple form of occurrence and co-occurrence of their printed forms in Section~\ref{Collapsing}. Our derivation of the emergence, interference and context equations  below  are however general.

Let us calculate the probability $\mu(AB)$ with which the combined concept $AB$, in the pre-measurement state represented by 
$|\psi'_{AB}\rangle = {N|\psi_{AB}\rangle \over \|N|\psi_{AB}\rangle\|}$, is considered to be meaning connected to a `story about $X$'. For this, we define the probabilities: 
\begin{equation}
p_A=\langle\psi_A | N |\psi_A\rangle,\quad p_B=\langle\psi_B | N |\psi_B\rangle,
\label{pApB}
\end{equation}
for the states represented by $|\psi_A\rangle$ and $|\psi_B\rangle$ to be eigenstates of the context $N$, respectively. According to (\ref{muAandmuB}), we have $\langle\psi_A | NM^s N|\psi_A\rangle = p_A\,\mu(A)$ and $\langle\psi_B | NM^s N|\psi_B\rangle =p_B\, \mu(B)$, so that the average $\mu(AB) = \langle\psi'_{AB} | M^s |\psi'_{AB}\rangle$ is given by: 
\begin{eqnarray}
 \mu(AB) &=& {\langle\psi_{AB} | N^\dagger M^s N|\psi_{AB}\rangle\over \|N|\psi_{AB}\rangle\|^2} = {\langle\psi_{AB} | NM^s N|\psi_{AB}\rangle\over \langle\psi_{AB} | N |\psi_{AB}\rangle}\nonumber \\
&=&{\langle\psi_A | NM^s N|\psi_A\rangle +\langle\psi_B | NM^s N|\psi_B\rangle + 2 \Re\, \langle\psi_A | NM^s N|\psi_B\rangle \over \langle\psi_A | N |\psi_A\rangle +\langle\psi_B | N |\psi_B\rangle + 2 \Re\, \langle\psi_A | N |\psi_B\rangle}\nonumber\\
&=&{p_A\,\mu(A) +p_B\, \mu(B) + 2 \Re\, \langle\psi_A | NM^s N|\psi_B\rangle \over p_A +p_B + 2 \Re\, \langle\psi_A | N |\psi_B\rangle},
\label{muAB}
\end{eqnarray}
where for the third term of the third equality in (\ref{muAB}), we have used: 
\begin{eqnarray}
\lefteqn{\langle\psi_A | NM^s N|\psi_B\rangle + \langle\psi_B | NM^s N|\psi_A\rangle =}\nonumber\\
&=& \langle\psi_A | NM^s N|\psi_B\rangle + \langle\psi_A | NM^s N|\psi_B\rangle^* =  2 \Re\, \langle\psi_A | NM^s N|\psi_B\rangle,
\end{eqnarray}
with the symbol $\Re$ denoting the real part of a complex number, and similarly for the term $2 \Re\, \langle\psi_A | N |\psi_B\rangle$ at the denominator.

We can observe that the first two terms at the numerator of (\ref{muAB}) can be interpreted as a `classical weighted average', whereas the third terms, at the numerator and denominator, are the so-called `interference contributions'. This becomes even more evident in the special situation where both states represented by $|\psi_A\rangle$ and $|\psi_B\rangle$ are already eigenstates of the context $N$, {\it i.e.}, $N|\psi_A\rangle=|\psi_A\rangle$ and $N|\psi_B\rangle=|\psi_B\rangle$, 
so that $p_A =p_B=1$, and (\ref{muAB}) simplifies to: 
\begin{equation}
\mu(AB)= {{1\over 2}[\mu(A) +\mu(B)] + \Re\, \langle\psi_A | M^s |\psi_B\rangle \over 1 + \Re\, \langle\psi_A |\psi_B\rangle},
\label{muAB-nocontext}
\end{equation}
with the weighted average now becoming a uniform one. Also, if $|\psi_A\rangle$ and $|\psi_B\rangle$ are assumed to be orthogonal vectors, (\ref{muAB-nocontext}) further simplifies to become: 
\begin{equation}
\mu(AB)= {1\over 2}[\mu(A) +\mu(B)] + \Re\, \langle\psi_A | M^s |\psi_B\rangle.
\label{muAB-nocontext-ortho}
\end{equation}

Emergent effects can generally be put in evidence as effects of `overextension' and `underextension' of the probabilities associated with the combined conceptual entity $AB$, where the terms refer to deviations with respect to classical disjunction and conjunction probabilities, respectively~\cite{h1988a,h1988b}. More precisely, one speaks of overextension if $\mu(AB) >\mu(A)$, or $\mu(AB) >\mu(B)$, and of `double overextension' if $\mu(AB) >\mu(A)$ and $\mu(AB) >\mu(B)$. On the other hand, one speaks of `underextension' if $\mu(AB) <\mu(A)$, or $\mu(AB) <\mu(B)$, and of `double underextension' if $\mu(AB) <\mu(A)$ and $\mu(AB) <\mu(B)$.

\section{Collapsing concepts onto words}
\label{Collapsing}

It is time to provide an example of the working of our quantum approach, when used to model the probabilities that are obtained by performing direct counts about specific words and combination of words appearing in the pages of the Web. Hence, in this section we will investigate in which way equation (\ref{muAB}), which we derived in all generality, copes with the simplified situation of co-occurrence. We proceed step by step, taking into account the operationality of our derivations, thus considering two special states for the two concepts $A$ and $B$.

To avoid possible confusions, let us stress again the difference between the notion of `state of a concept' and that of `story about a concept'. A story about a concept is a webpage, {\it i.e.}, a printed document. Webpages that are stories about a concept may or may not contain the word indicating such concept. Indeed, one can certainly write a text explaining what fruits are without ever mentioning the word ``fruits'' (using for instance, in replacement of the word ``fruits'', the term ``foods like apples, oranges, bananas, etc.,'' or the word ``fruits'' but written in a different language, like ``frutta,'' in Italian). Similarly, a text explaining what fruits are not, may contain the word ``fruits'' without actually being a story about fruits (not in the usual sense at least). On the other hand, a `state of $A$' expresses a condition which in general cannot be reduced to that of a story, as is clear that it can also be a superposition of ``stories about $A$,'' as expressed in (\ref{psiA}), and a superposition of stories is not anymore a story. 

In our example, we consider that $A$ and $B$ are described by states represented by the unit vectors
$|\psi_A\rangle$ and $|\psi_B\rangle$ that are superpositions only of stories ({\it i.e.}, states associated with webpages) that explicitly contain the words ``A'' and ``B.'' Also, we assume that these states express a `uniform meaning connection' towards all these stories. This means that we consider $|\psi_A\rangle$ and $|\psi_B\rangle$ to correspond to the `characteristic function states': 
\begin{equation}
|\chi_A\rangle={1\over \sqrt{n_A}} \sum_{j\in J_A} e^{i\alpha_j}|e_j\rangle,\quad |\chi_B\rangle={1\over \sqrt{n_B}} \sum_{j\in J_B} e^{i\beta_j}|e_j\rangle,
\label{characteristicAB}
\end{equation}
where $J_A$ and $J_B$ denote the sets of indexes associated with the webpages that are `stories about $A$' containing explicitly the word ``A,'' and `stories about $B$' containing explicitly the word ``B,'' respectively, 
which are of course subsets of the sets of indexes $I_A$ and $I_B$ associated with the stories about $A$ and $B$, not necessarily always manifestly containing these words. Here 
$n_A$ and $n_B$ are the numbers of these pages, {\it i.e.}, $|J_A|=n_A$ and $|J_B|=n_B$. Note that $|\chi_A\rangle$ is designated as a `characteristic function state' because it is a state (\ref{psiA}) such that its coefficients $a_j$ are of the form $a_j={1\over \sqrt{n_A}}\chi_{J_A}(j)$, where $\chi_{J_A}(j)$ is the characteristic (or indicator) function of the set $J_A$, {\it i.e.}, a function having value 1 for all $j\in J_A$, and 0 otherwise (and same for $|\chi_B\rangle$). So, the above is the operational way of expressing the notion of occurrence of the word ``A'' and occurrence of the word ``B'' within our general formalism. 

As a further simplification, we also consider that 
measurements are only about asking 
to tell `manifest stories about a concept', explicitly containing the word associated with said concept. Accordingly, we introduce the projection operator $M^{w}=\sum_{i\in J_X} |e_i\rangle\langle e_i|$, onto the set of vectors representing states that are generated by the webpages manifestly containing the word ``X,'' hence the superscript ``$w$'' that stands here for ``word.'' 
And this  is the operational way of expressing the notion of occurrence of the word ``X'' within our general formalism.

\subsection{Interference effects}
\label{no-context}

In addition to the above simplifications, let us also consider the situation where there are no context effects, {\it i.e.}, $N=\mathbb{I}$, or alternatively that the two states represented by $|\chi_A\rangle$ and $|\chi_B\rangle$ would be already eigenstates of $N$. Then $\mu(A) =\langle\chi_A | M^w |\chi_A\rangle$ and $\mu(B) =\langle\chi_B | M^w |\chi_B\rangle$, and according to (\ref{characteristicAB}), we have:
\begin{eqnarray}
M^w |\chi_A\rangle&=&\left(\sum_{j\in J_X} |e_j\rangle\langle e_j| \right)\left({1\over \sqrt{n_A}} \sum_{k\in J_A} e^{i\alpha_k}|e_k\rangle\right)\nonumber\\
&=& {1\over \sqrt{n_A}}\sum_{j\in J_X} \sum_{k\in J_A}e^{i\alpha_k}|e_j\rangle\underbrace{\langle e_j|e_k\rangle}_{\delta_{jk}}={1\over \sqrt{n_A}}\sum_{j\in J_{A,X}}e^{i\alpha_j}|e_j\rangle,
\label{Mwchi}
\end{eqnarray}
where $J_{A,X}$ denotes the sets of indexes associated with the webpages containing explicitly the words ``A'' and ``X.'' Therefore: 
\begin{eqnarray}
\mu(A)&=&\left(\langle\chi_A |\right)\left(M^w |\chi_A\rangle\right) = \left({1\over \sqrt{n_A}}\sum_{j\in J_A}  e^{-i\alpha_j}\langle e_j|\right)\left({1\over \sqrt{n_A}} \sum_{k\in J_{A,X}} e^{i\alpha_k}|e_k\rangle\right)\nonumber\\
&=& {1\over n_A}\sum_{j\in J_A} \sum_{k\in J_{A,X}}e^{i(\alpha_k-\alpha_j)}\underbrace{\langle e_j|e_k\rangle}_{\delta_{jk}}= {1\over n_A}\sum_{k\in J_{A,X}}\! 1,
\label{individualprob01}
\end{eqnarray}
and similarly for $\mu(B)$. So, if $n_{A,X}$ is the number of webpages containing both terms ``A'' and ``X,'' {\it i.e.}, $|J_{A,X}|=n_{A,X}$, and $n_{B,X}$ is the number of webpages containing both terms ``B'' and ``X,'' {\it i.e.}, $|J_{B,X}|=n_{B,X}$, we have for the individual probabilities:
\begin{equation}
\mu(A)=\langle\chi_A | M^w |\chi_A\rangle={n_{A,X}\over n_A},\quad \mu(B)=\langle\chi_B | M^w |\chi_B\rangle={n_{B,X}\over n_B}.
\label{individualprob}
\end{equation}
Thus, $\mu(A)$ and $\mu(B)$ can be simply interpreted as the probabilities of randomly selecting a webpage containing the term ``X,'' among those containing the terms ``A'' and ``B'', respectively. Hence, we have expressed within our general formalism the co-occurrence of the words ``X'' and ``A'' and the co-occurrence of the words ``X'' and ``B''.

According to (\ref{muAB-nocontext}), the probability $\mu(AB)$ for the combined concept $AB$ is:
\begin{eqnarray}
\mu(AB) &=& {{1\over 2}[{n_{A,X}\over n_A} +{n_{B,X}\over n_B}] + \Re\, \langle\chi_A | M^w |\chi_B\rangle \over 1 + \Re\, \langle\chi_A |\chi_B\rangle}\nonumber \\ 
&=& {{1\over 2}[{n_{A,X}\over n_A} +{n_{B,X}\over n_B}]+\sum_{j\in J_{AB,X}} {\cos(\beta_j-\alpha_j) \over \sqrt{n_An_B}}\over 1+ \sum_{j\in J_{AB}}{\cos(\beta_j-\alpha_j) \over \sqrt{n_An_B}}},
\label{muAB-nocontext-bis}
\end{eqnarray}
where for the second equality, we used (\ref{Mwchi}) and the fact that 
\begin{eqnarray}
\lefteqn{\Re\, \langle\chi_A | M^w |\chi_B\rangle = \Re\, \left({1\over \sqrt{n_A}} \sum_{k\in J_A} e^{-i\alpha_k}\langle e_k|\right)\left({1\over \sqrt{n_B}}\sum_{j\in J_{B,X}}e^{i\beta_j}|e_j\rangle\right)}\nonumber\\
&=&  {1 \over \sqrt{n_An_B}}\sum_{k\in J_A}\sum_{j\in J_{B,X}}\underbrace{\Re\, e^{i(\beta_j-\alpha_k)}}_{\cos(\beta_j-\alpha_k)}\underbrace{\langle e_k|e_j\rangle}_{\delta_{kj}} = \sum_{j\in J_{AB,X}} {\cos(\beta_j-\alpha_j)\over \sqrt{n_An_B}},
\end{eqnarray}
where $J_{AB,X}$ is the set of indexes associated with the webpages containing explicitly the words ``A,'' ``B'' and ``X,'' and similarly for the cosine term at the denominator of (\ref{muAB-nocontext-bis}), with $J_{AB}$ the set of indexes associated with the webpages containing explicitly the words ``A'' and ``B.''

We can see explicitly in (\ref{muAB-nocontext-bis}) the role played by the phases $\alpha_j$ and $\beta_j$, characterizing the two states represented by $|\chi_A\rangle$ and $|\chi_B\rangle$: by varying them, we do not alter the individual probabilities (\ref{individualprob}), but can extend the values taken by $\mu(AB)$ within a given interval of values. More precisely, by considering all possible phase differences $\beta_j-\alpha_j$, it becomes possible to generate an entire `interference interval', allowing to account for the overextension and underextension effects associated with the combination of the two concepts $A$ and $B$, which can make $X$ more or less meaning connected to $AB$, in comparison with the concepts $A$ and $B$ taken separately. This means that we have shown that, due to interference, there is place for the co-occurrence of ``A'' ``B'' and ``X'' to be independent of what is revealed in  the 
physical Web for 
the co-occurrence of ``A'' and ``X'' and the co-occurrence of ``B'' and ``X'' apart, expressing 
that these three situations of co-occurrence are linked to each other within the QWeb of meaning and not within the physical Web.

Let us analyze the width of this `interference interval'. For this, we observe that $J_{AB} =J_{AB,X}\cup J_{AB,X'}$, where $J_{AB,X'}$ denotes the set of indexes associated with the webpages containing the two terms ``A'' and ``B,'' but not ``X,'' and we have $|J_{AB,X}|=n_{AB,X}$, $|J_{AB,X'}|=n_{AB,X'}$, and $n_{AB}=n_{AB,X}+n_{AB,X'}$.  Thus, we can rewrite (\ref{muAB-nocontext-bis}) as follows:
\begin{equation}
\mu(AB)= {{1\over 2}[{n_{A,X}\over n_A} +{n_{B,X}\over n_B}]+\sum_{j\in J_{AB,X}} {\cos(\beta_j-\alpha_j) \over \sqrt{n_An_B}}\over 1+ \sum_{j\in J_{AB,X}}{\cos(\beta_j-\alpha_j) \over \sqrt{n_An_B}}+\sum_{j\in J_{AB,X'}}{\cos(\beta_j-\alpha_j) \over \sqrt{n_An_B}}}.
\label{muAB-nocontext-tris}
\end{equation}
It is then not difficult to verify that the maximum value $\mu^{\rm max}(AB)$ [minimum value $\mu^{\rm min}(AB)$] for $\mu(AB)$ is reached when the cosines terms at the numerator are all equal to $+1$ ($-1$), and the cosines terms at the denominator, for $j\in J_{AB,X'}$, are all equal to $-1$ ($+1$). Thus, the interference interval $I(AB) =[\mu^{\rm min}(AB),\mu^{\rm max}(AB)]$, describing the admissible values for $\mu(AB)$, is: 
\begin{equation}
I(AB)= \left[{{1 \over 2}[{n_{A,X} \over n_A}+{n_{B,X} \over n_B}]-{n_{AB,X} \over \sqrt{n_An_B}} \over 1- {n_{AB,X}-n_{AB,X'}\over \sqrt{n_An_B}}},{{1 \over 2}[{n_{A,X} \over n_A}+{n_{B,X} \over n_B}]+{n_{AB,X} \over \sqrt{n_An_B}} \over 1+ {n_{AB,X}-n_{AB,X'}\over \sqrt{n_An_B}}}\right].
\label{interval-int}
\end{equation}

Now, if we remain in the ambit of the ``manifest word'' approximation, and hence keep our focus on occurrence and co-occurrence, $\mu(AB)$ should be equal to the ratio ${n_{AB,X}\over n_{AB}}$, {\it i.e.}, to the probability of randomly selecting a webpage containing the term ``X,'' among those containing both terms ``A'' and ``B.'' Therefore, `interference effects' (with no `context effects') allow to model this situation only if ${n_{AB,X}\over n_{AB}}\in I(AB)$.

For numerous data, this will be the case, but interference effects are certainly not sufficient to model any possible situation, and hence express any type of `meaning dynamics with respect to occurrence and 
co-occurrence' taking place within the QWeb. Indeed, although $I(AB)$ can stretch up to the maximum interval $[0,1]$ (this is the case for instance if $A=B$, as it is easy to verify), in many situations the value ${n_{AB,X}\over n_{AB}}$ might lie outside of $I(AB)$. Just to give a simple example, consider the case where $n_{AB,X}=n_{AB}$. Then ${n_{AB,X}\over n_{AB}}=1$, and to model this situation we must have $\mu^{\rm max}(AB)=1$. However, it is easy to check on (\ref{interval-int}), considering that $n_{AB,X'}=0$, that this can be the case only if ${1 \over 2}[{n_{A,X} \over n_A}+{n_{B,X} \over n_B}]$ is equal to $1$, which of course will generally not be the case.

\subsection{Interference plus context effects}\label{context and interference}

We consider now the situation where also context effects apply, and prove that this allows all situations to be modeled, {\it i.e.}, by a suitable choice of the phases, $\mu(AB)$ can span the entire interval $[0,1]$. This means that if we allow context and interference effects to play a role on the QWeb, the data collected on the `physical Web' about occurrence and co-occurrence of words in a constellation as specified above (and consequently the QWeb's meaning dynamics giving rise to them) can always be modeled by our complex Hilbert space model of the QWeb. 

To make the situation more easy from a calculation perspective, we consider the 
simplifying assumption that $N$ is compatible with $M^w$, {\it i.e.}, $NM^w=M^wN$, from which it follows that $NM^w$ is also an orthogonal projector, {\it i.e.}, a self-adjoint and idempotent operator:
\begin{eqnarray}
&&(NM^w)^\dagger ={M^w}^\dagger N^\dagger = M^wN=NM^w,\nonumber\\
&&(NM^w)^2=NM^wNM^w=N^2(M^w)^2=NM^w.
\end{eqnarray}
We can thus introduce the three projection operators $P_1=M^wN$, $P_2=(\mathbb{I} -M^w)N$ and $P_3=\mathbb{I}-N$, which are orthogonal to each other:
\begin{eqnarray}
&&P_1P_2 = M^wN(\mathbb{I} -M^w)N = M^wN^2 - (M^wN)^2 = 0,\nonumber\\
&&P_1P_3 = M^wN(\mathbb{I}-N)=M^wN-M^wN^2=0,\nonumber\\
&&P_2P_3 =(\mathbb{I} -M^w)N(\mathbb{I}-N)=(\mathbb{I} -M^w)(N-N^2)=0.
\end{eqnarray}
We can thus write the Hilbert space ${\cal H}$ as the direct sum of the three orthogonal subspaces: ${\cal H}={\cal H}_1 \oplus {\cal H}_2\oplus {\cal H}_3$, where ${\cal H}_1= P_1 {\cal H}$, ${\cal H}_2= P_2 {\cal H}$ and ${\cal H}_3= P_3 {\cal H}$, and consequently, we can write $|\psi_A\rangle$ and $|\psi_B\rangle$ as the linear combinations:
\begin{eqnarray}
|\psi_A\rangle&=&ae^{i\alpha}|e\rangle+a'e^{i\alpha'}|e'\rangle+ a'' e^{i\alpha''} |e''\rangle, \\
|\psi_B\rangle&=&be^{i\beta}|f\rangle+b'e^{i\beta'}|f'\rangle+b'' e^{i\beta''} |f''\rangle, 
\end{eqnarray}
where $|e\rangle,|f\rangle\in {\cal H}_1$, $|e'\rangle,|f'\rangle\in {\cal H}_2$, $|e''\rangle,|f''\rangle\in {\cal H}_3$, and all vectors are unit vectors. We thus have $\langle\psi_A | NM^wN |\psi_A\rangle = \|P_1 |\psi_A\rangle\|^2=a^2$, and similarly $\langle\psi_B | NM^wN |\psi_B\rangle = \|P_1 |\psi_B\rangle\|^2=b^2$, so that:
\begin{equation}
p_A\mu(A) =\langle\psi_A | NM^wN |\psi_A\rangle = a^2, \quad p_B\mu(B) =\langle\psi_B | NM^wN |\psi_B\rangle  = b^2.
\end{equation}
Setting $\langle e|f\rangle = c\,e^{i\gamma}$ and $\phi =\gamma +\beta-\alpha$, we have: 
\begin{eqnarray}
&&\Re\, \langle\psi_A | NM^wN |\psi_B\rangle = \Re\, (\langle\psi_A | P_1)(P_1 |\psi_B\rangle) = \Re\, \left(\langle e|ae^{-i\alpha}\right)\left(be^{i\beta}|f\rangle\right)\nonumber\\
&&\quad\quad = ab\, \Re\, e^{i(\beta-\alpha)} \langle e|f\rangle = abc\,\Re\, e^{i(\gamma+\beta-\alpha)}= abc \cos\phi.
\end{eqnarray}
Also, setting $\langle e'|f'\rangle = c'e^{i\gamma'}$ and $\phi' =\beta'-\alpha'+\gamma'$, and considering that $\mathbb{I}=M^w + (\mathbb{I}- M^w)$, so that $N=[M^w + (\mathbb{I}- M^w)]N= P_1 +P_2$, we have:
\begin{equation}
\Re\, \langle\psi_A | N |\psi_B\rangle = \Re\, \langle\psi_A | P_1 |\psi_B\rangle +\Re\, \langle\psi_A |P_2 |\psi_B\rangle = abc \cos\phi+a'b'c'\cos\phi'.
\end{equation}
From (\ref{pApB}), we also obtain:  
\begin{equation}
p_A = \langle\psi_A | N |\psi_A\rangle = \langle\psi_A | P_1 |\psi_A\rangle + \langle\psi_A |P_2 |\psi_A\rangle = a^2 + {a'}^2
\end{equation}  
and similarly $p_B = b^2 + {b'}^2$, so that 
\begin{equation}
{a'}^2 = p_A-a^2 = p_A[1-\mu(A)],\quad {b'}^2 = p_B-b^2 = p_B[1-\mu(B)].
\end{equation}
Setting $\bar{\mu}(A) =1-\mu(A)$ and $\bar{\mu}(B) =1-\mu(B)$, we can now rewrite (\ref{muAB}) as:
\begin{equation}
\mu(AB) ={p_A\,\mu(A) +p_B\,\mu(B) + 2\sqrt{p_Ap_B}\sqrt{\mu(A)\mu(B)} c \cos\phi \over p_A+p_B+ 2\sqrt{p_Ap_B}[\sqrt{\mu(A)\mu(B)} c \cos\phi+\sqrt{\bar{\mu}(A)\bar{\mu}(B)} c' \cos\phi']}.
\label{muAB-bis}
\end{equation}

To relate the above expression with the webpages' counts, and hence express our simplified situation focused on occurrence and co-occurrence of words on the Web, we consider again the hypothesis that states are represented by characteristic functions, {\it i.e.}, by uniform superpositions of stories that explicitly contain the words ``A'' and ``B,'' respectively. More precisely, we assume that $|\psi'_A\rangle = |\chi_A\rangle$ and $|\psi'_B\rangle = |\chi_B\rangle $, so that:\footnote{Different from the previous `only interference effects situation', the vectors that we now assume to be represented by characteristic functions are those obtained following the action of the context $N$. This of course will not be true in general, and should only be considered as a rough approximation meant to illustrate that our approach can easily handle the probabilities calculated by performing webpages' counts (see Sec.~\ref{Numerical}).} $\mu(A) =\langle\psi'_A | M^w |\psi'_A\rangle = {n_{A,X}\over n_A}$ and $\mu(B) =\langle\psi'_B | M^w |\psi'_B\rangle = {n_{B,X}\over n_B}$. Hence, we then 
have again $\mu(AB)$ which is equal to ${n_{AB,X}\over n_{AB}}$, and the model will allow us to fit the data if we can find $p_A, p_B, c, c' \in [0,1]$ and $\phi, \phi' \in [0, 2\pi]$, such that the following equality holds: 
\begin{equation}
{n_{AB,X}\over n_{AB}} ={p_A\,{n_{A,X}\over n_A} +p_B\,{n_{B,X}\over n_B} + 2\sqrt{p_Ap_B}\sqrt{{n_{A,X}n_{B,X}\over n_An_B}} c \cos\phi \over p_A+p_B+ 2\sqrt{p_Ap_B}\left[\sqrt{{n_{A,X}n_{B,X}\over n_An_B}} c \cos\phi+\sqrt{{n_{A,X'}n_{B,X'}\over n_An_B}} c' \cos\phi'\right]}.
\label{muAB-tris}
\end{equation}

Eq. (\ref{muAB-tris}) will always be satisfied, since it is possible to show that (\ref{muAB-bis}) can deliver all values between $0$ and $1$. Consider the limit case $\mu(AB) =0$. Then, the numerator of (\ref{muAB-bis}) has to vanish. If for instance we choose $c=1$ and $\phi=\pi$, this means that we must have $[\sqrt{p_A\,\mu(A)}-\sqrt{p_B\,\mu(B)}]^2=0$, which is clearly satisfied if ${p_A\over p_B}={\mu(B)\over \mu(A)}$. For the case $\mu(AB) =1$, if we choose $c'=1$ and $\phi'=\pi$, (\ref{muAB-bis}) now gives 
the condition: $[\sqrt{p_A\,\bar{\mu}(A)}-\sqrt{p_B\,\bar{\mu}(B)}]^2=0$, which is clearly satisfied if ${p_A\over p_B}={\bar{\mu}(B)\over \bar{\mu}(A)}$.

What about all the intermediate values, between $0$ and $1$? If we set $\phi=\phi'={\pi \over 2}$, then (\ref{muAB-bis}) becomes: 
\begin{equation}
\mu(AB) ={p_A\over p_A+p_B}\,\mu(A) +{p_B\over p_A+p_B}\,\mu(B),
\label{muAB-4}
\end{equation}
which is a convex combination of $\mu(A)$ and $\mu(B)$, hence, by varying $p_A$ and $p_B$, all values in the interval $[\min(\mu(A),\mu(B)), \max(\mu(A),\mu(B))]$ can be reached. This allows to describe all possible non-double overextension and underextension effects, and since $\phi=\phi'={\pi \over 2}$ is a no-interference condition, this also means that context effects are sufficient alone to model situations of single overextension and single underextension. On the other hand, they are no more sufficient to model alone double overextension and double underextension effects, which require the relative phases $\phi$ or $\phi'$ to take values different from ${\pi \over 2}$.

To see that also the values in the `double underextension' and `double overextension' intervals $[0,\min(\mu(A),\mu(B))]$ and $[\max(\mu(A),\mu(B)),1]$ can be obtained, one needs to study the behavior of $\mu(AB)=\mu(AB;x,x')$ as a function of the two variables $(x,x')=(\cos\phi, \cos\phi')$. Let us consider first the case of the `double underextension' interval. We know that $\mu(AB;0,0)$ is given by (\ref{muAB-4}), so, we only need to show that, by varying $x$ and $x'$, we can reach the zero value (for suitable choices of $p_A$ and $p_B$). Now, for a given $x$, $\mu(AB;x,x')$ monotonically decreases as $x'$ increases. Thus, we only need to consider $\mu(AB;x,1)$. By studying the sign of $\partial_x\mu(AB;x,1)$, one sees that $\mu(AB;x,1)$ monotonically increases with $x$, so that the minimum corresponds to $\mu(AB;-1,1)$, which is equal to zero if $c=1$ and ${p_A\over p_B}={\mu(B)\over \mu(A)}$. Similarly, for the `double overextension' interval, we need to consider $\mu(AB;x,-1)$, and since $\mu(AB;x,-1)$ also monotonically increases with $x$, the maximum corresponds to $\mu(AB;1,-1)$, which is equal to $1$ if $c'=1$ and ${p_A\over p_B}={\bar{\mu}(B)\over \bar{\mu}(A)}$.

Thus, we have shown that for arbitrary $\mu(A)$, $\mu(B)$ and $\mu(AB)$, a quantum representation that combines context and interference effects always exists, which can faithfully model the experimental data.

\subsection{Numerical examples}\label{Numerical}
In this section, we provide a few examples of counts that we have performed on the Web by using Google as a search engine. Data reveal effects of both over- and under-extension, which in some cases can be modeled using either context effects or interference effects, taken separately, but in others require their combined action, to obtain a modeling of the data. The examples are inspired by Hamptons' celebrated psychological experiments, where overextension and underextension effects in membership judgments were evidenced for the first time \cite{h1988a,h1988b}.

It is important to mention that Google's counts are here meant to just illustrate our analysis and the importance of considering both context and interference effects as a way of modeling emergence of meaning due to concept combinations. This in particular because these counts are far from being a precise measure of the actual number of existing webpages containing specific words. However, the fact that a search engine does not provide accurate counts is also to be interpreted as a `context effect', taken into consideration in our model by means of the projection $N$. Anyhow, in order to improve the logical consistency of our counts, we have always performed ``three terms searches'' (also because the way Google does its searches can also depend on the number of words used, as this can force the engine to dig deeper, thus producing additional hits), so that instead of directly searching for $n_A$, we have deduced this number by calculating the sum: $n_A=n_{AB,X}+n_{AB',X}+n_{AB,X'}+n_{AB',X'}$. Similarly, $n_B=n_{AB,X}+n_{A'B,X}+n_{AB,X'}+n_{A'B,X'}$, $n_{AB}=n_{AB,X}+n_{AB,X'}$, $n_{A,X}=n_{AB,X}+n_{AB',X}$ and $n_{B,X}=n_{A,B,X}+n_{A',B,X}$.

We consider the situation where $A$ = {\it Fruits} and $B$ = {\it Vegetables}. With a search carried out on November 2016, we obtained $n_A=3.78 \cdot 10^{8}$, $n_B=3.57 \cdot 10^{8}$ and $n_{AB}=1.15 \cdot 10^{8}$. Then, we took for $X$ eight of the exemplars considered by Hampton: {\it Apple, Parsley, Yam, Elderberry, Olive, Raisin, Almond} and {\it Lentils}. For each of them, we did searches for the corresponding words,  to obtain the webpages' numbers $n_{A,X}$, $n_{B,X}$ and $n_{AB,X}$, then we deduced the numbers $n_{A,X'}=n_{A}-n_{A,X}$, $n_{B,X'}=n_{B}-n_{B,X}$ and $n_{AB,X'}=n_{AB}-n_{AB,X}$. In this way, we could calculate the probabilities $\mu(A)={n_{A,X}\over n_A}$, $\mu(B)={n_{B,X}\over n_B}$ and $\mu(AB)={n_{AB,X}\over n_{AB}}$, as well as the minimum and maximum values $\mu^{\rm min}$ and $\mu^{\rm max}$ of the interference interval (\ref{interval-int}).

The obtained values are reported in Table~\ref{table1}. As we can see, the first half of the exemplars that we considered produced values of ${n_{AB,X}\over n_{AB}}$ that can be modeled using only `interference effects', whereas the second half of them produced values of ${n_{AB,X}\over n_{AB}}$ that lie outside of the interference interval (\ref{interval-int}). Note also that only the data of the two exemplars {\it Parsley} and {\it Raisin} are contained in the interval $[\min({n_{A,X}\over n_A},{n_{B,X}\over n_B}), \max({n_{A,X}\over n_A},{n_{B,X}\over n_B})]$, and therefore could also be modeled using only `context effects'. Therefore, we already see in these examples that `interference and context effects' are jointly needed to faithfully account for the experimental data.

\begin{table}
\begin{center}
\begin{tabular}{||c|c|c|c|c|c||} 
\hline
X & ${n_{A,X}\over n_A}$ & ${n_{B,X}\over n_B}$ & ${n_{A,B,X}\over n_{A,B}}$ & $\mu^{\rm min}$ & $\mu^{\rm max}$\\ 
\hline\hline
{\footnotesize Apple} & $1.66\cdot 10^{-1}$ & $2.36\cdot 10^{-1}$ & $2.71\cdot 10^{-1}$ & $1.02\cdot 10^{-1}$ & $3.34\cdot 10^{-1}$\\ 
{\footnotesize Parsley} & $1.21\cdot 10^{-2}$ & $4.52\cdot 10^{-2}$ & $3.19\cdot 10^{-2}$ & $1.31\cdot 10^{-2}$ & $5.95\cdot 10^{-2}$\\
{\footnotesize Yam} & $2.88\cdot 10^{-3}$ & $3.48\cdot 10^{-3}$ & $4.76\cdot 10^{-3}$ & $1.15\cdot 10^{-3}$ & $7.30\cdot 10^{-3}$\\
{\footnotesize Elderberry} & $2.16\cdot 10^{-3}$ & $3.95\cdot 10^{-3}$ & $4.57\cdot 10^{-3}$ & $1.25\cdot 10^{-3}$ & $6.49\cdot 10^{-3}$\\
\hline\hline
{\footnotesize Olive} & $5.22\cdot 10^{-2}$ & $2.13\cdot 10^{-1}$ & ${2.90\cdot 10^{-1}}$ & $6.56\cdot 10^{-2}$ & $2.12\cdot 10^{-1}$\\ 
{\footnotesize Raisin} & $3.49\cdot 10^{-2}$ & $3.83\cdot 10^{-2}$ & ${1.04\cdot 10^{-1}}$ & $1.45\cdot 10^{-3}$ & $9.69\cdot 10^{-2}$\\ 
{\footnotesize Almond} & $9.01\cdot 10^{-2}$ & $1.10\cdot 10^{-1}$ & ${2.55\cdot 10^{-1}}$ & $6.21\cdot 10^{-3}$ & $2.35\cdot 10^{-1}$\\ 
{\footnotesize Lentils} & $1.42\cdot 10^{-2}$ & $1.69\cdot 10^{-2}$ & ${4.39\cdot 10^{-2}}$ & $1.38\cdot 10^{-3}$ & $4.10\cdot 10^{-2}$\\ 
\hline
\end{tabular}
\end{center}
\caption {Individual and joint probabilities for the words ``A = Fruits'' and ``B = Vegetables'', with respect to the exemplars ``X = Apple, Parsley, Yam, Elderberry, Olive, Raisin, Almond and Lentils.'' For ``Apple, Parsley, Yam, Elderberry'' data can be modeled using only `interference effects', {\it i.e.}, formula (\ref{muAB-nocontext-tris}), whereas for ``Olive, Raisin, Almond, Lentils'' the values of ${n_{AB,X}\over n_{AB}}$ lie outside of the `interference interval' $[\mu^{\rm min},\mu^{\rm max}]$.}
\label{table1}
\end{table}

Consider ``X = Olive.'' If we choose $p_A=p_B=c=c'=0.5$, we find that the `interference plus context' interval $[\mu(AB;-1,1),\mu(AB;1,-1)]$ is equal to $[5.78\cdot 10^{-2}, 2.98\cdot 10^{-1}]$, which clearly contains the value $2.90\cdot 10^{-1}$. For ``X = Raisin,'' if we take $p_A=0.2$, $p_B=0.8$, $c=0.3$ and $c'=0.8$, we find the interval $[1.79\cdot 10^{-2}, 1.18\cdot 10^{-1}]$, which contains $1.04\cdot 10^{-1}$. For ``X = Almond,'' we can choose $p_A=p_B=0.5$, $c=c'=0.6$, and the `interference plus context' interval is $[2.73\cdot 10^{-2}, 3.08\cdot 10^{-1}]$, which contains $2.55\cdot 10^{-1}$. Finally, for ``X = Lentils,'' choosing $p_A=p_B=c=c'=0.5$, we get $[5.25\cdot 10^{-3}, 4.52\cdot 10^{-2}]$, which contains $4.39\cdot 10^{-2}$.

\section{Discussion}\label{Conclusion}

We emphasized in this article  the importance of distinguishing the Web, made of actual (printed) webpages, from the meaning (conceptual) entity we have called the QWeb, our approach being a first tentative to explicitly model the latter, not the former. To do so, we have used a Hilbertian model and the associated Born rule, that is, the language of standard quantum theory. 
However, more general models could also be considered, generalizing the standard quantum formalism, like the recently derived GTR-model~\cite{asdb2015b,ass2016b}. This means that the ``Q'' in ``QWeb,'' which refers to the `quantum structure' of the mathematical model used to describe the `meaning entity associated with the Web', needs not to be understood in the limited sense of the standard quantum formalism. 

A valuable model of the QWeb must also allow one to account for the correlations that can be observed in the `Web of printed words'. Indeed, the relative frequency of co-occurrence of certain words in the webpages depends on the meaning connections between the concepts associated with these words. In that respect, we made a distinction between the notions of `state of a concept', `story about a concept' and `story about a concept manifestly containing the word denoting the concept'. We also assumed that an interrogation will generally be associated with a deterministic `context effect', which we modeled by means of an orthogonal projection operator.  In addition to these `context effects', we considered the possibility of modeling the emergence of meanings, when concepts are combined, by means of the quantum superposition principle, following the same modeling strategy that has proven successful in quantum cognition. 

More specifically, in Sec.~\ref{Collapsing}, we examined the special situation where the two concepts $A$ and $B$ are described by the characteristic function states given in (\ref{characteristicAB}). By doing so, and by also asking  
to only tell stories containing the explicit words of the concepts under investigation, we have derived the interference interval (\ref{interval-int}), which however cannot model all possible data. As we explained, one way to cope with this problem is to assume that context effects also play a role. But this is certainly not the only way to possibly further stretch the interval of values that can be modeled. Other possibilities, which we plan to explore in future research, are for instance: considering (i)
states whose meaning connections are not necessarily uniform, although still localized within the sets $J_A$ and $J_B$, or (ii) step function states extending beyond the `manifest word subspaces', 
for instance of the form: $|\psi_A^\epsilon\rangle= a\, |\chi_A\rangle + \bar{a}\, |\bar{\chi}_A\rangle$, where $|\bar{\chi}_A\rangle={1\over\sqrt{n-n_A}} \sum_{j\notin J_A}e^{i\alpha_j}|e_j\rangle$, and $|a|^2 + |\bar{a}|^2 =1$.

About our explicit numerical example of Sec.~\ref{Numerical}, the advantage of working with Google, or other search engines, is that we have in this way a very large collection of pages, with a lot of redundancies, which is convenient for obtaining relevant statistics, but then we have to use a very unreliable instrument to perform the counts. On the other hand, by using a dedicated software that can deliver exact counts, we would have the complementary problem of having to deal with a much smaller collection of documents, so that biases could now be introduced because of the way such collection would be constructed. 

Another remark is in order, which will help us to further clarify the scope of the present work, and avoid possible misunderstandings. One may object that, by considering a characteristic function state vector $|\chi_{AB}\rangle = \sum_{j\in J_{AB}}e^{i\delta_j}|e_j\rangle$, we can easily model all possible data, in the sense that, without even considering context effects, we would automatically have ${n_{AB,X}\over n_{AB}}=\langle \chi_{AB}|M^w|\chi_{AB}\rangle$, so that no phases and interference effects would be actually needed. This is correct, but would completely miss the point of our work. As we have already pointed out, our goal is modeling not the `Web of printed words' but, rather, the QWeb conceptual entity. In other words, we are primarily interested in describing and understanding the hidden structure of the Web, understood as a `complex meaning entity', and this independently of what we are today able of actually measure in practical terms. Our approach is meant to be a foundational one, not an {\it ad hoc} one.

Consider again the situation of the double-slit experiment in physics. One could also object that modeling the experiment by using a complex state vector to represent the superposition of the states for the single-slit situations would be a waste of time, as is clear that we can easily construct three different real functions to describe the distributions of impacts on the final screen: one function $f_A(x)$ for the situation where the $A$-slit is open and the $B$-slit is closed, a second real function $f_B(x)$ for the situation where the $B$-slit is open and the $A$-slit is closed, and a third real function $f_{A,B}(x)$ for the combined situation where both slits are open. Of course, these three functions would be perfectly able to do the job of describing what is observed in the three measurements, without for this the need of introducing superposition and interference effects, or complex numbers. However, they would not provide any explanation about how the fringe pattern can form on the screen. Also, they would not provide any clue about what would be measured if some of the parameters describing the experimental arrangement were varied, like for instance the distance between the two slits, or the energy of the incoming beam. 

{\it Mutatis mutandis}, the situation is the same with our modeling of the QWeb. Our example with the characteristic functions was meant to show that our understanding of how meaning is created when concepts are combined is consistent with what so far we can measure in experiments, like the idealized one we have considered, viewing the QWeb as an entity that can  
be subjected to `tell a story measurements'. 
These measurements certainly allow one to capture part of the existing meaning connections between the different concepts forming the quantum-like QWeb entity, but certainly we still lack today the sufficient knowledge, and technical tools, to fully describe all these connections. Our approach is in fact a first tentative to set up a possible stage for such an ambitious undertaking. 

An important distinguishing feature of our ``stage setting'' is the observation that one can speak of the meaning content carried by the Web in a way that is independent of the human minds that have created it, in the sense that we do not need human minds to find the traces that meaning can leave in the different physical (printed, or stored in memory) webpages. For example, these traces can be associated with the co-occurrence of certain words and combinations of words, in the different pages, which is something also a computer can easily detect, without the help from a human mind. It is indeed so that the co-occurrence of words is in no way related to the physical properties of the printed webpages, and is fully determined by the meaning carried by these webpages. In that sense, our example of Sect.~\ref{Collapsing} served a twofold purpose: that of illustrating our approach and testing its consistency, but also that of showing that meaning can ``stick out'' from the webpages (the written documents) -- for example in the case of co-occurring words -- in ways that can be accessed without requiring the intervention of a human mind. 

Of course, the meaning extending out of the webpages is not necessarily the entire meaning contained in the QWeb, in the same way that in quantum physics `collapsed states' of a measurement do not contain all the information about the pre-measurement, non-collapsed state. To be able to reconstruct the QWeb state, and therefore capture its entire meaning content, one should proceed in the same way as is done in so-called `quantum state tomography' \cite{dariano2004}, which is the process of reconstructing the complete initial state through a series of different measurements, characterized by different bases that need to be informationally complete. 

In the present article, we have only considered one of these bases, associated with the webpages, and at the moment it is still unclear what kind of additional QWeb measurements could be conceived. For instance, if the webpages' outcomes are considered to be the equivalent of a `position measurement', with each webpage representing a snapshot of (possibly a portion of) a ``space of printed or memorized documents and words,'' what would be the equivalent of a 
`momentum measurement'? Can such measurement be related, for instance, to the `impact' that a webpage can have, when it is read by a human mind?

We will not provide any tentative answer to these rather subtle questions, which we are motivated investigating in future works, apart observing that the `impact' concepts can have on one another can also be measured in terms of the interference phenomena they are able to produce, when they combine, which means that momentum-like measurements are certainly related to the values of the phases appearing in the complex exponential factors characterizing the concepts' states. 

Let us also note that if a `meaning content'  can be associated with the QWeb, in principle in an objective way, processes of meaning creation will also result from the interactions of  human minds with that content, which are highly contextual and subjective. This means that when addressing the problem of devising advanced information retrieval systems, based on meaning, not only the modeling of the QWeb will play a role, but also the modeling of how human minds interact with the latter. In this article, we have only considered measurements consisting in 
telling stories. However, when a human mind interacts with  
one of the outcome-states of these measurements,  
this can also be described as a measurement context, the outcomes of which are for instance the different possible understandings that such human mind can have of that given QWeb  
outcome-state. 
This is what is typically investigated in cognitive psychology experiments, and can also be effectively modeled using the mathematical formalism of quantum theory (quantum cognition; see for example: \cite{a2009a,pb2009,k2010,bpft2011,bb2012,hk2013,pb2013,wbap2013,abgs2013,ags2013,ast2014,asdb2015b}). Cognitive psychology experiments usually involve a number of different participants, all submitted to the same conceptual situation (the same state), the outcomes of which are  counted to deduce  their relative frequencies, interpreted as the probabilities for the different outcomes. So, we have to do with two different quantum modeling: one describing the QWeb meaning entity, and the other describing how different QWeb 
outcome-states 
are processed by collections of different human minds, and of course both modeling will play a role in information retrieval systems.

Having said this, we would like now to conclude on a more philosophical, and in a sense also more speculative, note. Our considerations about the printed webpages and the fact that we cannot expect them to contain the entire meaning of the QWeb, also apply, \emph{mutatis mutandis}, to the hardware structures of classical computers, as is clear that also these structures do not allow for superpositions. Hence, when we think of our brain as a classical machine, we should ask how it is possible for this machine to capture the full content of the associated mind. One possibility is of course that of exploring the numerous dualistic views, where cognitive and material aspects are considered to be fundamentally distinct  \cite{Robinson2017}.  However, if one wants to remain within the confines of a materialistic paradigm, then the states of the physical brain need to be similar in structure to the mental states. And since mental states appear to be characterized by the possibility of creating superpositions, as only in this way new meanings can be created, one needs to be able to describe human (but also animal) brains not as classical entities, but as quantum ones. 

So, when adopting a materialistic view on the human mind, the default analogy should be that of a `quantum computer', and not of a `classical computer'. To put it differently, because of our parochial viewpoint on the physical reality (resulting from our long evolution on the surface of this planet, in our very special bodies \cite{as2015}), we only focused on the classical structure associated with the different physical entities, and forgot (or never realized) that such structure is just the spatial tip of a much deeper non-spatial structure, in the same way that the Web is just the tip of a much deeper and more complex quantum-like entity, which we called the QWeb. From this oversight (or error), another followed: that of splitting our personal reality according to a dualistic mind-body view, consequence of the fact that we are not used to associate cognitive properties to physical objects, and consequently also to our physical bodies and brains. 

When we look for a way to extract meaning from a corpus of printed documents, willy-nilly we touch at the essence of the mind-body problem, and by connecting printed words with concepts we automatically open ourselves to the possibility of connecting objects with conceptual entities, viewing both as aspects of a unique reality. This is precisely the approach that one of us has boldly taken in recent years, when proposing a new interpretation of quantum theory, called the `conceptuality interpretation' \cite{aerts2009,aerts2010a,aerts2010b,aerts2013,aerts2014},  where 
quantum entities like electrons, protons, etc., are precisely viewed as entities having a conceptual nature, playing the role of communication vehicles between material entities composed of ordinary matter, which function for them as memory structures. This is the same approach we have taken in this work, where the Web is viewed as a memory formed by printed pages, which however interacts, or better communicates, with the different conceptual entities at a more abstract level. This communication, in certain contexts, can produce traces in the memory structure, which are precisely the words that make up the stories contained in the different pages, from which part of the meaning that has generated them can be reconstructed. 

We believe that it is only when this broader conceptuality approach is adopted in relation to documental entities, and an operational-realistic quantum  modeling of its conceptual structure is attempted (as we did for the QWeb model we sketched in this article, using interference and context effects), that a deeper understanding of how meaning can leave its traces in documents can be accessed, possibly leading to the development of more context-sensitive and semantic-oriented information retrieval models. This is a challenge that we are willing to take with our group, in the years to come.

\end{document}